\documentclass[letter,12pt]{article}
\usepackage{amsmath,graphicx,multicol}
\usepackage{amsfonts}
\usepackage{color}
\usepackage{bbm}

\usepackage{amssymb,parskip}
\usepackage{algorithm}
\usepackage{booktabs}
\usepackage[margin=1.1in]{geometry}
\usepackage{hyperref}
\usepackage{tabularx}
\usepackage[utf8]{inputenc}
\usepackage[english]{babel}
\usepackage{amsthm}
\usepackage{subcaption}
\usepackage[font={footnotesize}]{caption}
\usepackage{url}
\usepackage{mathtools}
\usepackage{comment}
\usepackage{float}
\usepackage[toc,page]{appendix}
\usepackage{amssymb}
\usepackage{enumitem}
\usepackage{fancyhdr}
\newtheorem{definition}{Definition}[]
\newtheorem{proposition}{Proposition}[]
\newtheorem{theorem}{Theorem}[]

\newtheorem{assumption}{Assumption}

\pdfoutput=1

\usepackage{tikz}
\usepackage{listings}
\usepackage{algpseudocode}
\definecolor{codegreen}{rgb}{0,0.6,0}
\definecolor{codeblack}{rgb}{0,0,0}
\definecolor{codepurple}{rgb}{0.58,0,0.82}
\definecolor{backcolour}{rgb}{1.0,1.0,01.0}
\lstdefinestyle{mystyle}{
    backgroundcolor=\color{backcolour},
    commentstyle=\color{codegreen},
    keywordstyle=\color{magenta},
    numberstyle=\tiny\color{codeblack},
    stringstyle=\color{codepurple},
    basicstyle=\ttfamily\footnotesize,
    breakatwhitespace=false,         
    breaklines=true,                 
    captionpos=b,                    
    keepspaces=true,                 
    numbers=left,                    
    numbersep=5pt,                  
    showspaces=false,                
    showstringspaces=false,
    showtabs=false,                  
    tabsize=2
}
\usepackage{ dsfont }
\usepackage{enumitem}
\usepackage{biblatex} %Imports biblatex package
\usepackage{booktabs}
\usepackage{multirow}
\usepackage{graphicx}
\title{\vspace{0mm}\textbf{Ortho-ODE: Enhancing Robustness and  of Neural ODEs against Adversarial Attacks }}
\author{Vishal Purohit}

\begin{document}
\pagenumbering{arabic}

\maketitle
\tableofcontents
%%%%%%%%%%%%%%%%%%%%%%%%%%%%%%%%%%%%%%%%%%%%%%%%%%%%%%%%%%%%%
%%%%%%%%%%%%%%%%%%%%%%%%%%%%%%%%%%%%%%%%%%%%%%%%%%%%%%%%%%%%%
\begin{abstract}%   <- trailing '%' for backward compatibility of .sty file
Neural Ordinary Differential Equations (NODEs) proposed in the influential work \cite{3}, probed the usage of numerical solvers to solve the differential equation characterized by a Neural Network(NN), therefore initiating a new paradigm of deep learning models with infinite depth. NODEs were designed to tackle the irregular time series problem. However, NODEs have demonstrated robustness against various noises and adversarial attacks. This paper is about the natural robustness of NODEs and examines the cause behind such surprising behavior. We show that by controlling the Lipschitz constant of the ODE dynamics the robustness can be significantly improved. We derive our approach from Grownwall’s inequality. Further, we draw parallels between contractivity theory and Grownwall’s inequality. Experimentally we corroborate the enhanced robustness on numerous datasets - MNIST, CIFAR-10, and CIFAR 100. We also present the impact of adaptive and non-adaptive solvers on the robustness of NODEs.	
\end{abstract}
\section{Introduction}
Deep Learning (DL) has revolutionized and impacted diverse fields of science. It has found successful applications in high-level vision tasks like - image classification, object detection, and image segmentation, and low-level tasks like image super-resolution, deburring, etc. Despite a plethora of applications, deep learning algorithms suffer from fundamental problems that limit their application to critical fields like medical imaging, security, and surveillance. But \cite{5} found that most of the existing state-of-the-art neural networks are easily fooled by adversarial examples that are generated using tiny perturbations to the input images. The Inputs corrupted with imperceptible perturbations can easily fool many vanilla deep neural networks (DNNs) into misclassifying them and degrading their performance. A new subfield of deep learning, adversarial attacks \cite{4}, is dedicated to designing such imperceptible perturbations to data and defenses for such attacks. Recently, \cite{1} \cite{2} have applied Neural Ordinary Differential Equations (NODEs) \cite{3} to defend against adversarial attacks. Some works \cite{1} explored the natural robustness of NODEs against adversarial attacks, both, empirically and theoretically. The work \cite{1} made some interesting observations and provided the reasoning behind such surprising properties of NODEs. NODEs were introduced to tackle the irregular time series problem but their surprising robustness against attacks has piqued the interest of a lot of researchers. Though NODEs are robust against adversarial attacks, they still suffer from poor performance on specific attacks, specifically gradient-free attacks. Practically, it is impossible to defend against every adversarial attack out in the wild. Meanwhile, a more important question to be asked is - why NODEs are robust against some adversarial attacks, and how to improve their robustness?

So far many techniques have been introduced to tackle adversarial attacks. Probably one of the most successful techniques is \textit{adversarial training} introduced in works \cite{6} \cite{7}. In adversarial training, the adversarial examples are simulated in each iteration of the model and used as a training set in the next iteration of training. Using adversarial training is computationally expensive since in every iteration we need to generate adversarial examples and retrain the model on newly generated samples. In contrast, NODEs offer robustness naturally without the need for adversarial training which makes them attractive to computation-limited applications.

In this paper, our objective is to assess the effect of the Lipschitz constant of dynamics of NODE on the robustness of the model against adversarial attacks. To this end, we first propose to use orthogonal convolutional layers \cite{8} using Cayley transform to design the NN that signifies the dynamics of non-linear dynamical system. Encouraging orthogonality in neural networks has proven to yield several compelling benefits. Our work specifically uses two such benefits - preserving gradient norms and enforcing low Lipschitz constants. Controlling Lipschitz constants is nontrivial and has shown several benefits \cite{9}, \cite{10}, \cite{11} against perturbations in Convolutional Neural Networks (CNNs). Different from CNNs, NODEs are infinite depth resnets \cite{12}. Because of their infinite depth nature, we need to ensure that NODEs do not suffer from degraded activations due to exploding and vanishing gradients \cite{13}. Orthogonal convolutional layers using the Cayley transform ensure stable activation, preserving gradient norms and enforcing low Lipschitz constants. We call our method as \textit{Ortho-ODE} as in ODE with orthogonal convolutional layers. Our method is backed theoretically by Grownwall’s inequality.
Our contributions are:

\begin{itemize}
\item Our method proposes to use orthogonal convolutional layers to characterize the NN representing dynamics of ODE. Thus, enabling us to upper bound the Lipschitz constant of the dynamics making our model robust.
\item We theoretically justify that imposing orthogonality constraints on dynamics ensures the representations of adversarial and pure samples remain close. Therefore, increasing the classification accuracy of our method.
\item We test our method on multiple datasets and against many state-of-the-art robust NODE methods. We draw parallels between our method and contractivity theory to demonstrate that various theoretical pieces of evidence support our method.
\end{itemize}

\section{Related Work}

\subsection{Neural-ODEs}
NODEs were first introduced in the work by Chen et al. \cite{3} as continuous depth formulation of ResNet architecture. Many notable architects can be interpreted as different discretizations of the differential equations \cite{15}. Many of the subsequent works have followed up with an exploration of optimization issues and the expressivity of NODE. For example, in the work \cite{16} it was shown that the expressivity of the NODEs is limited due to the topology-preserving nature of NODEs. To overcome such issues \cite{16} presented augmentedODE to learn more complex functions. The work \cite{1} was the first to evaluate the adversarial robustness of NODEs theoretically as well as empirically.

Additionally, in \cite{17} it was shown that injecting noise could be beneficial to the stability of NODEs. Despite some appealing properties of NODEs, they are computationally expensive. Hence a recent study \cite{18} explored the efficient implementation of the adjoint training method.  The renewed interest in the marriage of dynamic systems and deep learning  has given rise to a plethora of works combining the theory of dynamical systems with NODEs.  In the original formulation of NODE, there was no depth or input-dependent modification of the dynamics. However, \cite{19} suggested using neural ODE whose dynamics would depend on the input.

\subsection{Adversarial Attacks} 
Adversarial examples are seen as threats to neural networks, especially in critical applications. Adversarial examples are fed to the neural network to modify their predictions to the desired prediction. As one of the initial applications of adversarial attacks dates back to work  \cite{20}, \cite{21} which modified the spellings of the words to fool the spam filters. But, the first adversarial attacks on computer vision models were introduced in work \cite{5} and \cite{22}. These two works established a new field of deep learning focusing on the design and defense of adversarial examples. It is puzzling to many researchers as to why neural networks are sensitive to imperceptible perturbations (adversarial attacks) in the image. The work \cite{23} proposed a localize the attack region to a small patch instead of adding noise to the whole image. Many such attacks have been formulated for speech processing \cite{24}, \cite{25}. Multiple works \cite{26},\cite{27} have proposed attacks designed specifically for adaptive models similar to NODEs.

Since the introduction of adversarial examples, many works have been proposed to refine the attacks and target various properties of the neural network. Broadly, adversarial attacks could be classified as - black-box and white-box attacks \cite{37}. Black box attacks are those, where the attacker does not have access to any knowledge about the model or its output. However, in white box attacks, the attacker has access to the gradient information, outputs, or model architecture. White box attacks are generally more effective because of the design of targeted attacks designed for a specific model using the available information. Two of the most famous attacks are FGSM \cite{38} and PGD \cite{38}, which is a white box attack with the goal of misclassification. PGD is a gradient-based attack where the attacker has access to gradients of the model during training. Since then many sophisticated attacks have been proposed. Autoattack \cite{38} is a suite of attacks carefully designed to do large-scale evaluations of the robustness of NNs.

\subsection{Adversarial Defense}
The adversarial defense literature is equally rich with the most famous being adversarial training \cite{22}, Bayesian adversarial training \cite{15}, and other regularization-based methods \cite{11}. Many variations of adversarial training approaches also have been proposed \cite{33}. Among various defense mechanisms,  \cite{28} was the first to work to improve the robustness using control theory and dynamic systems. Further, the use of Lyapunov-stable equilibrium for NODEs is investigated in Stable Neural ODE \cite{29}.

\section{Methodology}
In this section, we first go over some of the preliminaries and problem formulation for image classification using NODEs. We follow up with a detailed description of methods and theorems supporting our hypothesis. 
\subsection{Preliminaries on Neural ODE}
Under the neural ODE framework, we model the input and output as two states of a continuous dynamical system by approximating the system's dynamics with trainable layers. The Neural ODEs are endowed with intrinsic invertibility, yielding to a family of invertible models for solving inverse problems \cite{30}. The following equations characterize the relation between input and output:
\begin{equation}\label{eq1}
    \frac{dz(t)}{dt} = f_{\mathcal{W}}(z(t), t), z(0) = z_{in}  
\end{equation}
where $z_{out} = z(T)$ and $f_{\mathcal{W}} : \mathbb{R}^n \times [0, \infty) \rightarrow \mathbb{R}^{n}$ denotes the trainable layers that are parameterized by weights $\mathcal{W}$. We assume that $f_{\mathcal{W}}$ is continuous in $t$  and globally Lipschitz continuous in $z$. The input of neural ODE corresponds to the state at $t=0$ and output $z_{out}$ is associated with the state at some $T \in (0, \infty)$. Given the input $z_{in}$, the output $z_{out}$ at time $t$ can be calculated by solving ODE in Eq. \ref{eq1} Therefore, the solution of neural ODE can be represented as a $d-$dimensional function $\phi_{T}(.,.)$ i,e.
\begin{equation}
    z_{out} = z(T) = z(0) + \int_{0}^{T} f_{\mathcal{W}}(z(t), t) dt = \phi_{T}(z_{in}, f_{\mathcal{W}})
\end{equation}
It is quite easy to see that NODEs are the continuous analog of residual networks where the hidden layer of the resnet can be regarded as discrete-time difference equations $z(t+1) = z{t} + f_{\mathcal{W}}(z(t), t)$.

For classification, task NODEs cannot directly be applied to images. We need CNN layers before the NODE layer to extract the representation before passing it to NODE. Additionally, we need a set of fully connected layers post NODEs to classify the representation in various classes. The pre and post-CNN layers be represented by $f_{pre}(.)$ and $f_{post}(.)$. NODEs have a property that is formalized using the following proposition
\begin{proposition}(ODE integral curves do not intersect \cite{31}, \cite{32}) Let \(f\) be a function whose derivative exists in every point, then \(f\) is a continuous function. Let $z_1(t)$ and $z_2(t)$ be two solutions of the ODE with different initial conditions, i.e. $z_1(0) \neq z_2(0)$. $f_{\mathcal{W}}$ is continuous in t, and globally Lipschitz is continuous in z.
\end{proposition}
Then, it holds that $$z_1(t) \neq z_2(t)$$ for all $$t \in [0,\infty)$$

\subsection{Problem Formulation}
Let $\mathcal{P}_{data}$ be the probability distribution of the data over $\mathcal{X} \times \mathcal{Y}$, where $\mathcal{X}$ represent a set of input data points and $\mathcal{Y}$ represents the corresponding labels. Let $n^{th}$ pair of data be represented by pair $(x_n, y_n)$, where $x_{n} \in \mathbb{R}^{m \times n}$ represents an image of $m \times n$, width and height respectively and $y_n \in \mathbb{R}^C$. Here $C$ is the total number of classes in the training dataset. Further, we assume that both training and test data come from $\mathcal{P}_{data}$. We extract the features from input $x_n$ using CNN layers and process the output of NODE using an MLP layer. The whole processing pipeline is represented by the following functions -
\begin{center}
    \includegraphics[scale=0.55]{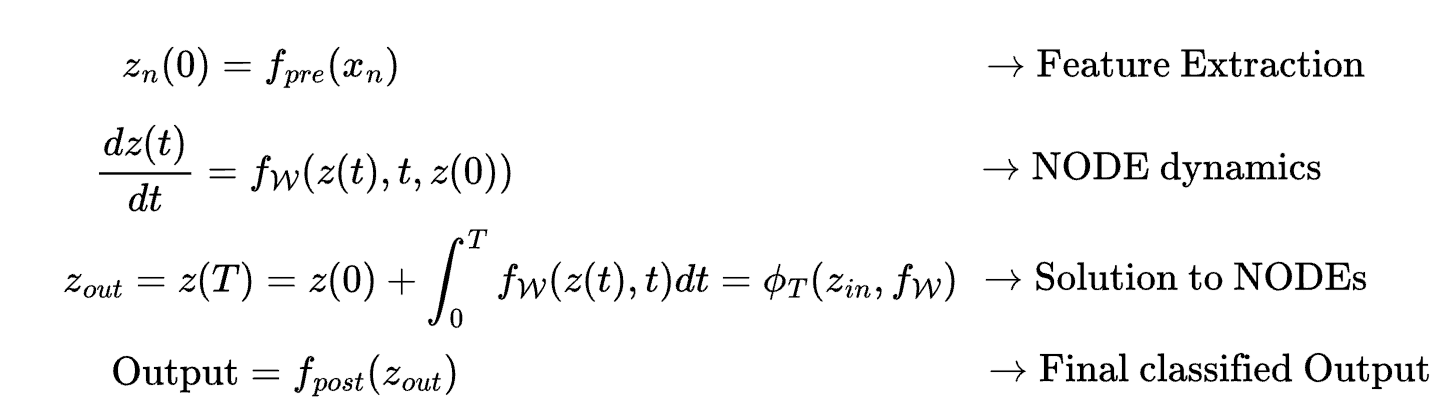}
\end{center}
During test time the adversarial samples are generated using PGD or FSGM be represented by $x_{adv}$. Our goal is to correctly classify the adversarial samples without adversarial training of NODE.
\subsection{The Grönwall–Bellman inequality}
 A matrix $A \in \mathbb{R}^{m \times n}$ is said to be an orthogonal matrix if $A^TA = I_{n}$, where $I_{n}$ is the identity matrix of $n \times n$ dimension. Matrix $A$ is said to be \textit{semi-orthogonal} if $A^TA = I_{n}$ or $AA^T = I_{n}$. If $m \geq n$, then $A$ is norm preserving, and if $m \leq n$ then the mapping is contractive. Alternatively, a matrix with all singular values is orthogonal too. One of the consequences of orthogonality is 1-Lipschitzness. The definition of Lipschitzness is as below  
\begin{definition}(Lipschitness) A function $f: \mathbb{R}^n \rightarrow \mathbb{R}^m$ is $\textit{L}$-Lipschitz w.r.t $l_2$ norm if and only if $\frac{||f(x) - f(x^{'})||_{2}}{||x - x^{'}||_2} \leq L$ $\forall x, x^{'} \in \mathbb{R}^n$. $\textit{L}$ is called lipschitz constant of $f$
\end{definition}
\begin{figure}[t]
    \centering
    \includegraphics[scale=0.35]{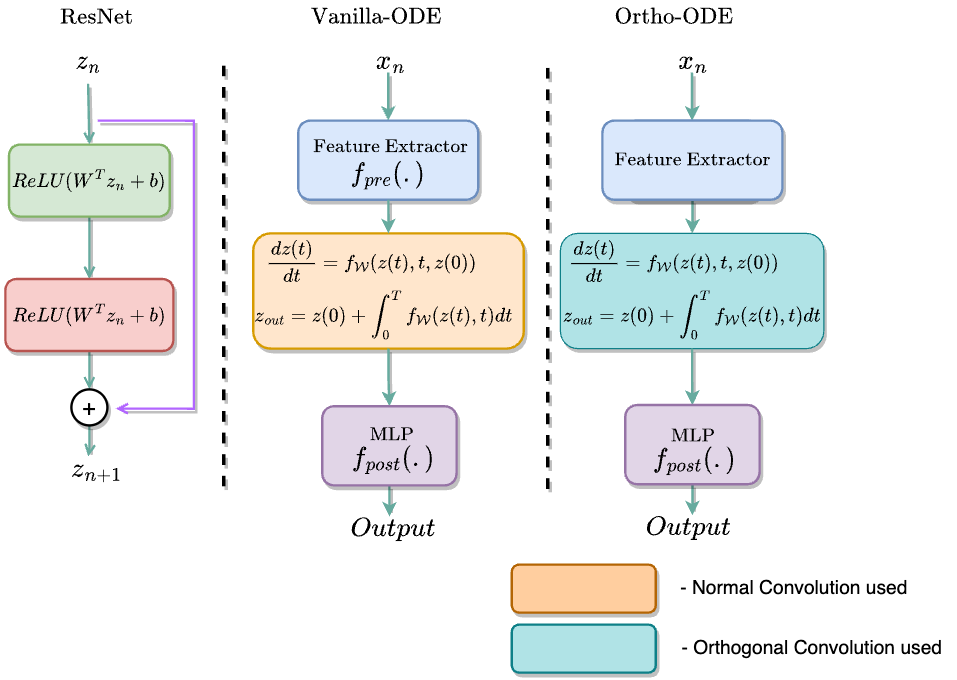}
    \caption{Architecture difference between Resnet, Vanilla ODE and Ortho-ODE}
    \label{fig:my_label}
\end{figure}

There are multiple works showcasing that bounding the Lipschitz constant of a neural network has guaranteed robustness against small perturbations in the input. Though the benefit of bounding the Lipschitz constant has been explored for neural networks the impact of bounding the Lipschitz constant on the dynamics of NODE is rarely investigated. According to  Grownall's inequality \cite{14}, when stated informally - The difference between two terminal states of NODE is bounded by the difference between the initial states times the exponential of the Lipschitz constant of the dynamics. In neural ODE, dynamics are represented by a neural network. Hence controlling the Lipschitz constant, one can control how close two outputs are though initial conditions are quite different. Formally the theorem is as below-
\begin{theorem}(Grönwall–Bellman inequality)\label{the1}
Let $U \subset \mathbb{R}^d$ be an open set. Let $f: U \times [0, T] \rightarrow \mathbb{R}^d$ be a continuous function and let $z_1$ and $z_2$ satisfy the Initial Value Problem (IVP) problems -
\begin{equation}
    \frac{dz_1(t)}{dt} = f(z_1(t_0),t), z_1(t_0) =  x_1
\end{equation}
\begin{equation}
    \frac{dz_2(t)}{dt} = f(z_2(t_0),t), z_1(t_0) =  x_2
\end{equation}
Let $C$ be a constant such that $C \geq 0$ such that $\forall t \in [0, T]$,
\begin{equation}
    ||f(z_2(t_0), t) - f(z_1(t_0), t)|| 
\leq C||z_2(t) - z_1(t)||
\end{equation}
Consequently, 
\begin{equation}
    ||z_2(t) - z_1(t)||
\leq ||x_2 - x_1|| . e^{Ct}
\end{equation}
Where $C$ is the Lipschitz constant of the dynamics.
\end{theorem}
The key to improving the robustness of the NODEs is to control the difference between the neighboring curves. From Theorem \ref{the1} we can bring two neighboring curves close together by bounding the Lipschitz constant of the dynamics of ODE. Directly bounding the Lipschitz constant is hard. Hence we resort to Cayley transform to impose orthogonality constraints on the dynamics of ODE. 
For our method to work, we make the following assumption -
\begin{assumption}(Representation closeness)
Let $z_{pure}$ represent the input representation of \textit{pure} sample and $z_{adv}$ be the input representation of an adversarial sample. Then we assume that $z_{adv}$ is in the $\epsilon$ neighborhood of $z_{pure}$. i.e,
\end{assumption}
\begin{equation}
    ||z_{pure} - z_{adv}|| \leq \epsilon
\end{equation}

\subsection{Connection between Grownwall's Inequality and Contraction theory}
Recently contraction theory has been employed in Neural Networks for multiple purposes. For example, \cite{33} explored the use of contractivity to improve the well-posedness and robustness of implicit neural networks, and analysis of Hopfield NNs \cite{34}. In \cite{39} author proposed Hamiltonian NODE which is contractive by design to improve the robustness. Singh et al. \cite{35} and Revay et al. \cite{36} utilize contraction theory to learn stabilizable nonlinear NN models from available data. Contractivity is the property of a dynamical system and it ensures that trajectories of the dynamical system converge to each other asymptotically. Formally contractivity of a dynamic system is defined below -
\begin{definition}(Contractivity) The dynamics of an ODE is contractive with contraction rate $\zeta > 0$ if 
\begin{equation}
    ||\hat{z}_t - z_{t}|| \leq e^{-\rho t} ||\hat{z_0} - z_0||, \hspace{10pt}\forall t \in [0, T]
\end{equation}
where $z_0, \hat{z}_0 \in \mathbb{R}^n$ are the initial conditions of IVP and $\hat{z}_t$ and $z_t$ are its solutions.
\end{definition}
Therefore we can say that a NODE is contractive, the Lipschitz constant between the input and the output is smaller than 1. Making NODEs globally contractive can significantly hamper expressive power. Our goal is to apply local contractivity efficiently to harness the natural robustness provided by locally contractive NODEs.

\subsection{Ortho-ODE}
A recent work by Torckman et al.\cite{8}  parameterized orthogonal convolutions using Cayley transform in a scalable and efficient way. The key idea behind the method in \cite{8} is that  multi-channel convolution in the Fourier domain reduces to a batch of matrix-vector products, and making each of those matrices orthogonal makes the convolution orthogonal. Since orthogonalization directly controls the Lipschitz constant, we propose to model the layers of the neural network describing the dynamics using orthogonal convolution layers instead of normal convolution layers. A comparison between ResNets, Vanilla ODE, and Ortho-ODE is shown in Figure.\ref{fig:my_label}. We briefly go over the Cayley transform and how it can be used to impose orthogonal constraints on convolution operation. 

Consider convolutional layer with stride 1 with $c_{in}$ representing the input channel and $c_{out}$  representing the output channels. Let the set of weights of the convolutional layer mapping from input to output is of shape $c_{out} \times c_{in} \times n \times n$, where $n \times n$ is the size of the convolutional kernel. The convolutions are easier to be analyzed when they are considered to be circular. The convolution is said to be circular when if the kernel goes out of the bounds of input, it wraps around to the other side of the input. We define $Conv(X)$ as the circular convolutional layer with weight tensor $W \in \mathbb{R} ^{c_{\text{out}} \times c_{\text{in}} \times n \times n}$ applied to an input tensor $X \in \mathbb{R} ^{b \times c_{\text{in}} \times h \times w}$. The resulting output tensor $Y = Conv(X) \in \mathbb{R}^{cout×n×n}$. One can view the convolutional operation as doubly block-circulant matrix $C \in \mathbb{R}^{c_{out} n^2 c_{in}n^2}$. Similarly, we denote by $Conv^{T}(X)$ the transpose of the same convolution.

The Cayley transform is a bĳection between skewsymmetric matrices $A$ and orthogonal matrices $Q$ without \textminus$1$ eigenvalues:
\begin{equation}Q = (I - A)(I + A)^{-1} \end{equation}

A matrix is said to be skew-symmetric if $A = -A^T$. The Cayley transform of such a skew-symmetric matrix is always orthogonal. Since convolutions are linear transformations we can the Cayley transform to convolutions. As described in \cite{8} While it is possible to construct the matrix $C$ corresponding to a convolution $Conv$ and apply the Cayley transform to it, this is highly inefficient in practice: Convolutions can be easily skew-symmetrized by computing $Conv(X) - Conv^{T}(X)$, but finding their inverse is challenging; instead, we manipulate convolutions in the Fourier domain, taking advantage of the convolution theorem and the efficiency of the fast Fourier transform.

Since we can construct multiple layers of CNN where each layer is orthogonal and we use such layers to construct the neural network that represents the dynamics of the NODE. As verified in \cite{8} using the clayey transform it is possible to maintain the orthogonalization constraint consistently.

\section{Experiments}
In this section, we first describe the datasets used to benchmark our algorithm followed by details of existing ODE and Non-ODE based benchmarks used for comparison. Further, we describe the training settings including model description, metric and numerical results. \\

\textbf{Datasets}
General datasets used for evaluating the robustness of NODEs include MNIST, CIFAR-10, and CIFAR-100. MNIST is the easiest dataset among others. CIFAR-10 and CIFAR-100 are difficult datasets used for the classification task. For each dataset, we have 60,000 training samples  and 10,000 test samples. MNIST and CIFAR-10 have 10 classes but CIFAR-100 has 100 classes.\\

\textbf{Benchmarks} Our method with orthogonal layers is dubbed as \textit{Ortho-ODE}. We compare Ortho-ODE with parameter-wise equivalent ResNet-10, Vanilla-ODE which uses normal convolutional layers, and TisODE \cite{1}\\

\textbf{Training Details}
All the methods are trained for 100 epochs with a learning rate of 0.01 and no weight decay. We ensure the number of parameters in all the architectures is almost similar. We augment each data with crop and rotation augmentation. We evaluate all the models in the absence and presence of adversarial samples. We use PGD and FGSM  adversarial attacks to assess the robustness of our method. Additionally, we use gaussian noise with different standard deviations to assess the robustness against non-target attacks.\\

\textbf{Metrics}
We present classification accuracy for each method on each dataset\\

\subsection{Numerical Results}
We evaluate our method against several benchmarks. We briefly describe the benchmarks used for comparison. We evaluate all the methods in two training configurations - one with adversarial attacks and another without adversarial attacks. Additionally, we also evaluate the performance in the presence of Gaussian Noise. ResNet10 is a normal CNN with residual connections and there is a total of 10 layers, making its number of parameters almost equivalent to parameters in other methods. As expected ResNet10 does not perform well when it comes to adversarial attacks. On both FGSM and PGD-based attacks, ResNet10 struggles to give any good accuracy. All the results are tabulated in Table \ref{tab:table_1}. We find that ResNet10 outperforms all the methods under no attack configuration. 

We evaluate our method against Neural ODE or Vanilla ODE where the convolutional layers used in neural network parameterize the dynamics are normal convolutional layers. We ensure that all channels and the total number of parameters remain almost the same as our method. The performance of Vanilla ODE is far better than ResNet10 under all attacks. However, under no attack configuration, ResNet10 dominates. 

Further, TisODE uses the time invariance property to ensure that in the solution of two slightly different initial conditions the final output from NODE almost remains the same. In our, evaluation TisODE performs best in most cases under different adversarial attacks. Compared to our method, underperforms in some cases but there is no clear evidence suggesting that our method completely outperforms TisODE.

\begin{table}[]
\resizebox{\columnwidth}{!}{%
\begin{tabular}{@{}cclc|cc|cc@{}}
\toprule
\multirow{2}{*}{Datasets} & \multirow{2}{*}{Methods} & \multirow{2}{*}{No Attacks} & Gaussian & \multicolumn{2}{c|}{FGSM} & \multicolumn{2}{c}{PGD} \\ \cmidrule(l){4-8} 
          &                        &                & $\sigma = 100$ & FSGM - 5/255   & FSGM - 8/255   & PGD - 0.2      & PGD -0.3       \\ \midrule
MNSIT     & ResNet10               & 99.15          & 98.75          & 28.20          & 16.07          & 32.67          & 0.0            \\
          & Vanilla ODE            & 97.84          & 97.63          & 49.10          & 34.78          & 64.89          & 13.02          \\
          & TisODE                 & 99.13          & 98.9           & 50.23          & 36.98          & 67.47          & 13.7           \\
          & Ortho-ODE (Our Method) & 99.14          & \textbf{99.10} & 49.32          & \textbf{37.52} & \textbf{67.86} & 11.56          \\ \midrule
CIFAR-10  & ResNet10               & 91.12          & 90.56          & 38.10          & 17.05          & 30.45          & 1.2            \\
          & Vanilla ODE            & 82.7           & 81.30          & 42.89          & 38.12          & 49.18          & 12.56          \\
          & TisODE                 & 85.30          & 84.12          & 44.23          & 37.46          & 50.34          & 13.1           \\
          & Ortho-ODE (Our Method) & \textbf{85.69} & 85.10          & 43.12          & 36.89          & \textbf{50.78} & 11.4           \\ \midrule
CIFAR-100 & ResNet10               & 68.57          & 66.89          & 18.57          & 14.54          & 17.13          & 0.0            \\
          & Vanilla ODE            & 52.91          & 56.21          & 47.67          & 37.67          & 21.89          & 11.12          \\
          & TisODE                 & 53.62          & 55.71          & 48.12          & 36.41          & 23.72          & 12.34          \\
          & Ortho-ODE (Our Method) & \textbf{53.64} & 52.45          & \textbf{49.45} & 35.32          & 21.81          & \textbf{12.56} \\ \bottomrule
\end{tabular}%
}
\caption{Robustness Results of our method compared against existing approaches on MNIST, CIFAR-10, and CIFAR-100}
\label{tab:table_1}
\end{table}
\section{Limitations \& Conclusion}
Apart from poor performance under adversarial attacks, our method inherits the disadvantages of the Cayley transform. Among all the methods shown in the results table, our method is the slowest due to the use of the Cayley transform which converts the signal to the Fourier domain using the Fast Fourier transform. As normal convolution would not require a such step, hence our method is the slowest. 

Apart from slow training and inference, the accuracy of our method needs to improve to be competitive with TisODE. Due to the orthogonality constraints on the convolutional layers, we sacrifice the expressivity of the model to some extent. The trade-off between expressivity and robustness is a common theme shared across multiple algorithms. 

In this work, we evaluated the use of Grownwall's inequality to improve the robustness of NODE. We constrained the Lipschitz constant of the neural network representing the dynamics using orthogonal constraints. We choose to use Cayley transform to constrain and impose the orthogonality requirement. We evaluate our method on multiple datasets and compared it against various benchmarks. Though our method does not outperform the benchmarks in all cases, our method still works and sometimes outperforms TisODE.
\section{Future Work}
In the future, we would like to further explore the connection between the bounding of the Lipschitz constant of NODE and the adversarial robustness. It is important to figure out a faster method to impose the orthogonality requirement in order to improve the training an inference speed of our method.

\printbibliography %Prints bibliography
\end{document}